\documentclass[11pt,a4paper]{article}
\usepackage[hyperref]{emnlp-ijcnlp-2019}
\usepackage{times}
\usepackage{latexsym}
\usepackage{url}

\usepackage{multirow}
\usepackage{mathrsfs}
\usepackage{amsfonts,amssymb}
\usepackage{amsmath}
\usepackage{booktabs}
\usepackage{graphicx}
\usepackage{subfigure}
\usepackage{diagbox}
\aclfinalcopy 

\title{Dually Interactive Matching Network for Personalized Response Selection in Retrieval-Based Chatbots}

\author{Jia-Chen Gu$^1$, Zhen-Hua Ling$^1$, Xiaodan Zhu$^2$, Quan Liu$^{1,3}$ \\
  $^1$National Engineering Laboratory for Speech and Language Information Processing, \\
      University of Science and Technology of China, Hefei, China \\
  $^2$ECE, Queen's University, Kingston, Canada \\
  $^3$State Key Laboratory of Cognitive Intelligence, iFLYTEK Research, Hefei, China \\
{\tt gujc@mail.ustc.edu.cn}, {\tt zhling@ustc.edu.cn}, \\ {\tt xiaodan.zhu@queensu.ca}, {\tt quanliu@ustc.edu.cn}
}

\date{}

\begin{document}
\maketitle
\begin{abstract}
  This paper proposes a dually interactive matching network (DIM) for presenting the personalities of dialogue agents in retrieval-based chatbots. This model develops from the  interactive matching network (IMN) which models the matching degree between a context composed of multiple utterances and a response candidate. Compared with previous \textit{persona} fusion approaches which enhance the representation of a context by calculating its similarity with a given persona, the DIM model adopts a dual matching architecture, which performs interactive matching between responses and contexts and between responses and personas respectively for ranking response candidates. Experimental results on PERSONA-CHAT dataset show that the DIM model outperforms its baseline model, i.e., IMN with \textit{persona} fusion, by a margin of 14.5\% and outperforms the current state-of-the-art model by a margin of 27.7\% in terms of top-1 accuracy $\textbf{hits}@1$.
\end{abstract}

\section{Introduction}

  Building a conversation system with intelligence is challenging. Response selection, which aims to select a potential response from a set of candidates given the context of a conversation, is an important technique to build retrieval-based chatbots \cite{DBLP:conf/acl/WuLCZDYZL18}. Many previous studies on single-turn \cite{DBLP:conf/emnlp/WangLLC13} or multi-turn response selection \cite{DBLP:conf/sigdial/LowePSP15,DBLP:conf/acl/WuLCZDYZL18,DBLP:conf/cikm/GuLL19} rank response candidates according to their semantic relevance with the given context.

  With the emergence and popular use of personal assistants such as Apple Siri, Google Now and Microsoft Cortana, the techniques of making personalized dialogues  has attracted much research attention in recent years \cite{DBLP:conf/acl/LiGBSGD16,DBLP:conf/acl/KielaWZDUS18,DBLP:conf/emnlp/MazareHRB18}. \citet{DBLP:conf/acl/KielaWZDUS18} constructed a PERSONA-CHAT dataset for building personalized dialogue agents, where each \textit{persona} was represented as multiple sentences of profile description. An example dialogue conditioned on given profiles from this dataset is given in Table \ref{tab1} for illustration.

  A persona fusion method for personalized response selection was also proposed by \citet{DBLP:conf/acl/KielaWZDUS18}.
  In this method, given a context and a persona composed of several profile sentences, the similarities between the context representation and all profile sentences are computed first using attention to get the persona representation.
  Then, the persona representation is applied to enhance the context representation by a simple concatenation or addition operation.
  Finally, the enhanced context representation is used to rank response candidates.
  This method has two main deficiencies.
  First, the context is treated as a whole for calculating its attention towards profile sentences. However, each context is composed of multiple utterances and these utterances may play different roles when matching different profile sentences.
  Second, the interactions between the persona and each response candidate are ignored when deriving the persona representation.

    \begin{table*}[!hbt]
     \small
     \centering
     \begin{tabular}{c|l|c|l}
      \toprule
       \multicolumn{2}{c|}{\textbf{Persona 1}} & \multicolumn{2}{c}{\textbf{Persona 2}} \\
      \midrule
       \multirow{5}{*}{Original} & I just bought a brand new house.   & \multirow{5}{*}{Original} &  I love to meet new people. \\
                                 & I like to dance at the club.       &                           &  I have a turtle named timothy. \\
                                 & I run a dog obedience school.      &                           &  My favorite sport is ultimate frisbee. \\
                                 & I have a big sweet tooth.          &                           &  My parents are living in bora bora. \\
                                 & I like taking and posting selkies. &                           &  Autumn is my favorite season. \\
      \midrule
       \multirow{5}{*}{Revised}  & I have purchased a home.                     & \multirow{5}{*}{Revised} & I like getting friends. \\
                                 & Just go dancing at the nightclub, it is fun! &                          & Reptiles make good pets. \\
                                 & I really enjoy animals.                      &                          & I love to run around and get out my energy. \\
                                 & I enjoy chocolate.                           &                          & My family lives on a island. \\
                                 & I pose for pictures and put them online.     &                          & I love watching the leaves change colors. \\
      \midrule
       \multicolumn{4}{l}{\textbf{Dialogue}} \\
       \multicolumn{4}{l}{Person 1: Hello, how are you doing tonight?} \\
       \multicolumn{4}{l}{Person 2: I am well an loving this interaction how are you?} \\
       \multicolumn{4}{l}{Person 1: I am great. I just got back from the club.} \\
       \multicolumn{4}{l}{Person 2: This is my favorite time of the year season wise.} \\
       \multicolumn{4}{l}{Person 1: I would rather eat chocolate cake during this season.} \\
       \multicolumn{4}{l}{Person 2: What club did you go to? Me an timothy watched tv.} \\
       \multicolumn{4}{l}{Person 1: I went to club chino. What show are you watching?} \\
       \multicolumn{4}{l}{Person 2: LOL oh okay kind of random.} \\
       \multicolumn{4}{l}{Person 1: I love those shows. I am really craving cake.} \\
       \multicolumn{4}{l}{Person 2: Why does that matter any? I went outdoors to play frisbee.} \\
       \multicolumn{4}{l}{Person 1: It matters because I have a sweet tooth.} \\
       \multicolumn{4}{l}{Person 2: So? LOL I want to meet my family at home in bora.} \\
       \multicolumn{4}{l}{Person 1: My family lives in alaska. It is freezing down there.} \\
       \multicolumn{4}{l}{Person 2: I bet it is oh I could not.} \\

      \bottomrule
      \end{tabular}
      \caption{An example dialogue from the PERSONA-CHAT dataset.}
      \label{tab1}
    \end{table*}

  In this paper, the interactive matching network (IMN) \cite{DBLP:conf/cikm/GuLL19} is adopted as the fundamental architecture to build our baseline and improved models for personalized response selection. The baseline model follows the persona fusion method proposed by \citet{DBLP:conf/acl/KielaWZDUS18} and two improved models are then proposed.
  First, an IMN-based persona fusion model with fine-grained context-persona interaction is designed. In this model, each utterance in a context, instead of the whole context, is used to calculate its similarity with each profile sentence in a persona.
  Second, a dually interactive matching network (DIM) is proposed by formulating the task of personalized response selection as a dual matching problem, i.e., finding a response that can properly match the given context and persona simultaneously. The DIM model calculates the interactions between the context and the response, and between the persona and the response in parallel, in order to derive the final matching feature for  response selection.

  We test our proposed methods on the PERSONA-CHAT dataset \cite{DBLP:conf/acl/KielaWZDUS18}. Results show that the IMN-based utterance-level persona fusion model and the DIM model can obtain a top-1 accuracy $\textbf{hits}@1$ improvement of 2.4\% and 14.5\%, respectively, over the baseline model, i.e., the IMN-based context-level persona fusion model. Finally, our proposed DIM model outperforms the current state-of-the-art model by a margin of 27.7\% in terms of top-1 accuracy $\textbf{hits}@1$ on the PERSONA-CHAT dataset.

  In summary, the contributions of this paper are three-fold.
  (1) An IMN-based fine-grained persona fusion model is designed in order to consider the utterance-level interactions between contexts and personas.
  (2) A dually interactive matching network (DIM) is proposed by formulating the task of personalized response selection as a dual matching problem, aiming to find a response that can properly match the given context and persona simultaneously.
  (3) Experimental results on the PERSONA-CHAT dataset demonstrate that our proposed models outperform the baseline and state-of-the-art models by large margins on the accuracy of response selection.

\section{Related Work}

  \subsection{Response Selection}

    Response selection is an important problem in building retrieval-based chatbots. Existing work on response selection can be categorized into single-turn \cite{DBLP:conf/emnlp/WangLLC13} and multi-turn dialogues \cite{DBLP:conf/sigdial/LowePSP15,DBLP:conf/acl/WuLCZDYZL18,DBLP:conf/cikm/GuLL19}. Early studies have been more on single-turn dialogues, considering only the last utterance of a context for response matching. More recently, the research focus has been shifted to multi-turn conversations, a more practical setup for real applications. \citet{DBLP:conf/acl/WuWXZL17} proposed the sequential matching network (SMN) which first matched the response with each context utterance and then accumulated the matching information by a recurrent neural network (RNN). \citet{DBLP:conf/acl/WuLCZDYZL18} proposed the deep attention matching network (DAM) to construct representations at different granularities with stacked self-attention. \citet{DBLP:conf/cikm/GuLL19} proposed the interactive matching network (IMN) to enhance the representations of the context and response at both the word-level and sentence-level, and to perform the bidirectional and global interactions between the context and response in order to derive the matching feature vector.

  \subsection{Persona for Chatbots}

    Chit-chat models suffer from a lack of a consistent personality as they are typically trained over many dialogues, each with different speakers, and a lack of explicit long-term memory as they are typically trained to produce an utterance given only a very recent dialogue history. \citet{DBLP:conf/acl/LiGBSGD16} proposed a persona-based neural conversation model to capture individual characteristics such as background information and speaking style. \citet{DBLP:conf/emnlp/MillerFDKBW16} proposed the key-value memory network, where the keys were dialogue histories, i.e., contexts, and the values were next dialogue utterances. \citet{DBLP:conf/acl/KielaWZDUS18} proposed the profile memory network by considering the dialogue history as input and then performing attention over the persona to be combined with the dialogue history. \citet{DBLP:conf/emnlp/MazareHRB18} proposed the fine-tuned persona-chat (FT-PC) model which first pretrained a model using a large-scale corpus with external knowledge and then fine-tuned it on the PERSONA-CHAT dataset.

    In general, all these methods adopted a context-level persona fusion strategy, which first obtained the embedding vector of a context and then computed the similarities between the whole context and each profile sentence to acquire the persona representation. However, such persona fusion is relatively too coarse. The utterance-level representations of contexts are not leveraged. The interactions between the persona and each response candidate are also ignored when deriving the persona representation.

\section{Task Definition}

  Given a dialogue dataset $\mathcal{D}$ with personas, an example of the dataset can be represented as $(c,p,r,y)$.
  Specifically,
  $c = \{u_1,u_2,...,u_{n_c}\}$ represents a context with $\{u_m\}_{m=1}^{n_c}$ as its utterances and $n_c$ as the utterance number.
  $p = \{p_1,p_2,...,p_{n_p}\}$ represents a persona with $\{p_n\}_{n=1}^{n_p}$ as its profile sentences and $n_p$ as the profile number.
  $r$ represents a response candidate.
  $y \in \{0,1\}$ denotes a label. $y=1$ indicates that $r$ is a proper response for $(c,p)$; otherwise, $y=0$.
  Our goal is to learn a matching model $g(c,p,r)$ from $\mathcal{D}$. For any context-persona-response triple $(c,p,r)$, $g(c,p,r)$ measures the matching degree between $(c,p)$ and $r$. A softmax output layer over all response candidates is adopted in this model. The model parameters are trained by minimizing a multi-class cross-entropy loss function on $\mathcal{D}$.

\section{IMN-Based Persona Fusion}

   \begin{figure}
    \centering
    \subfigure[Context-level persona fusion]{
    \includegraphics[width=7cm]{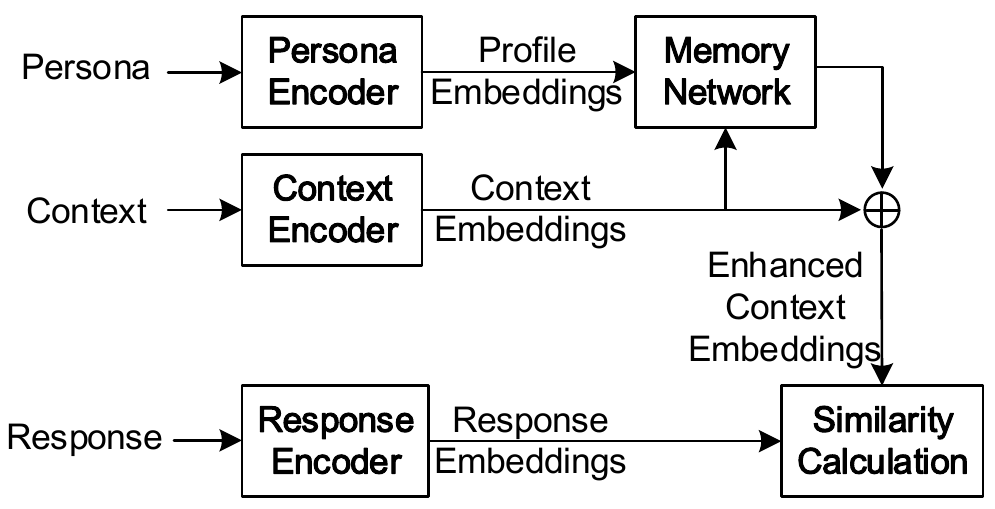}}
    \subfigure[Utterance-level persona fusion]{
    \includegraphics[width=7cm]{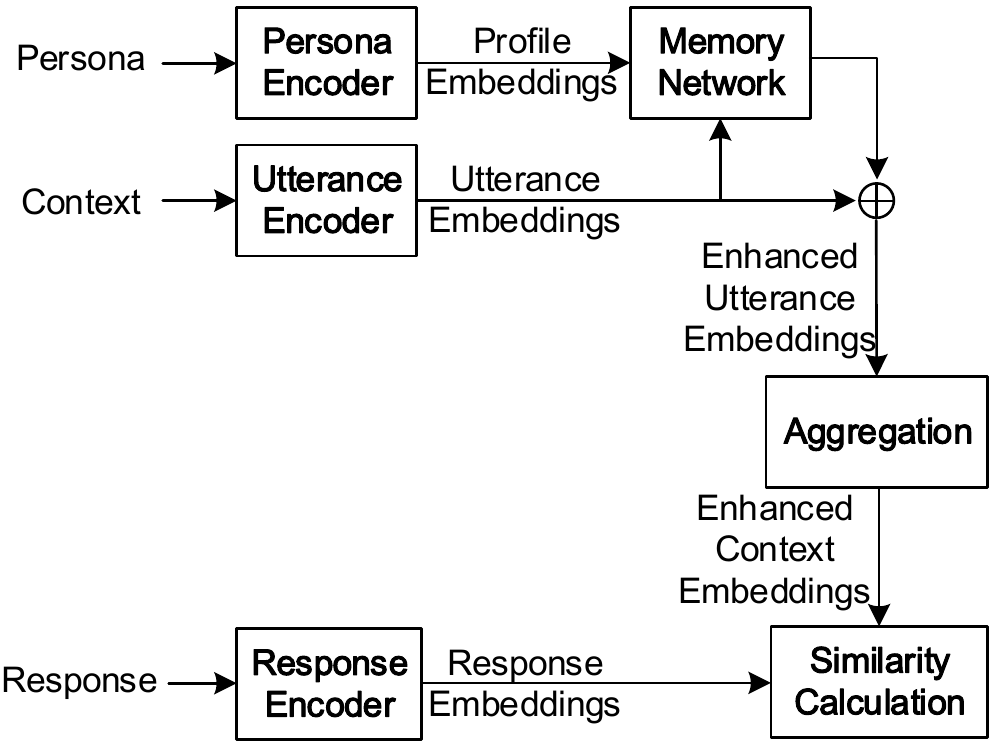}}
    \caption{Comparison of the model architectures for (a) context-level persona fusion and (b) utterance-level persona fusion.}
    \label{fig3}
   \end{figure}

   The model architecture used by previous methods with persona fusion \cite{DBLP:conf/acl/KielaWZDUS18,DBLP:conf/emnlp/MazareHRB18} is shown in Figure~\ref{fig3}(a). It first obtains the context representation and then computes the similarities between the whole context and each profile sentence in a persona. Attention weights are calculated for all profile sentences to obtain the persona representation. Finally, the persona representation is combined with the context representation through concatenation or addition operations.

   Formally, the representations of the whole context which is the concatenation of utterances, the context utterances, and the profile sentences are denoted as $\textbf{c}$, $\{\textbf{u}_m\}_{m=1}^{n_c}$ and $\{\textbf{p}_n\}_{n=1}^{n_p}$ respectively, where $\textbf{c}$, $\textbf{u}_m$ and $\textbf{p}_n \in \mathbb{R}^d$. In previous context-level persona fusion methods, the enhanced context representation $\textbf{c}^+$ fused with persona information is calculated as
    \begin{equation}
    \textbf{c}^+ =  \textbf{c} + \sum_{n} \textbf{Softmax}(\textbf{c} \cdot \textbf{p}_n) \textbf{p}_n.
    \label{equ2}
    \end{equation}
   Then, the similarity between  $\textbf{c}^+$ and the response representation are computed to get the matching degree of $(c,p,r)$.

   In this paper, we build our baseline model based on IMN \cite{DBLP:conf/cikm/GuLL19}.
   After the context and response embeddings are obtained in the IMN model, the context-level persona fusion architecture shown in Figure~\ref{fig3}(a) is applied to integrate persona information. All model parameters are estimated in an end-to-end manner. This baseline model is denoted as IMN$_{ctx}$ in this paper.

   Considering each context is composed of multiple utterances and these utterances may play different roles when matching different profile sentences, we propose to improve the baseline model by fusing the persona information at a fine-grained utterance-level as shown in Figure~\ref{fig3}(b). This model is denoted as IMN$_{utr}$ in this paper. First, the similarities between each context utterance and each profile sentence are computed and the enhanced representation $\textbf{u}_m^+$ of each context utterance is calculated as
   \begin{equation}
   \textbf{u}_m^+ =  \textbf{u}_m + \sum_{n} \textbf{Softmax}(\textbf{u}_m \cdot \textbf{p}_n) \textbf{p}_n.
   \end{equation}
   Then, these enhanced utterance representations are aggregated into the enhanced context representation as
    \begin{equation}
    \textbf{c}^+ = \textbf{Aggregation}( \{\textbf{u}_m^+\}_{m=1}^{n_c} ),
    \end{equation}
   where either RNN or attention-based aggregation \cite{DBLP:conf/cikm/GuLL19} can be employed.

\section{Dually Interactive Matching Network}

  \subsection{Model Overview}

   \begin{figure*}
    \centering
    \includegraphics[width=16cm]{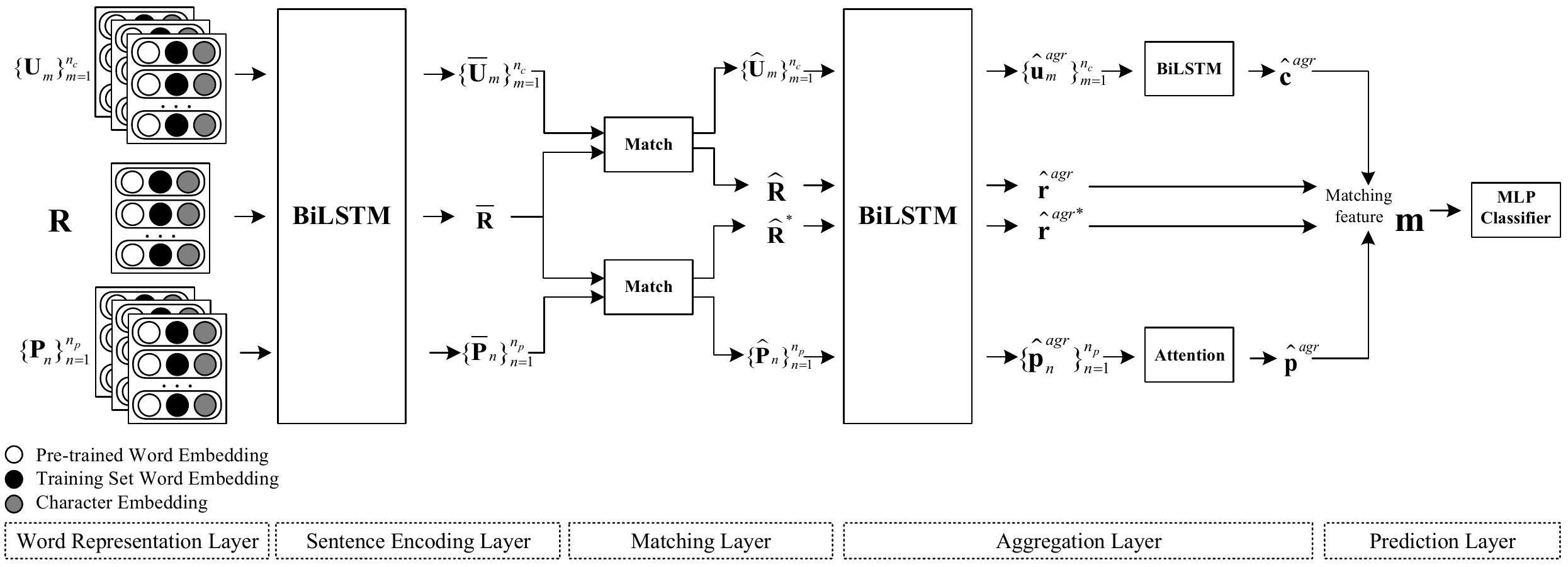}
    \caption{An overview of our proposed DIM model.}
    \label{fig1}
   \end{figure*}

   Previous studies on personalized response selection treat personas as supplementary information to enhance context representations by attention-based interaction. In this paper, we formulate the task of personalized response selection as a dual matching problem. The selected response is expected to properly match the given context and persona respectively. Here, personas are considered as equally important counterparts to contexts for ranking response candidates. The interactive matching between the context and response, and that between the persona and response constitute the dually interactive matching network (DIM).

   The DIM model is composed of five layers. Figure~\ref{fig1} shows an overview of the architecture. Details about each layer are provided in the following subsections.

  \subsection{Word Representation Layer}
   We follow the setting used in IMN \cite{DBLP:conf/cikm/GuLL19}, which constructs word representations by combining general pre-trained word embeddings, those estimated on the task-specific training set, as well as character-level embeddings, in order to deal with the out-of-vocabulary issue.

   Formally, embeddings of the \emph{m}-th utterance in a context, the \emph{n}-th profile sentence in a persona and a response candidate are denoted as $\textbf{U}_m = \{\textbf{u}_{m,i}\}_{i=1}^{l_{u_m}}$, $\textbf{P}_n = \{\textbf{p}_{n,j}\}_{j=1}^{l_{p_n}}$ and $\textbf{R} = \{\textbf{r}_k\}_{k=1}^{l_r}$ respectively, where $l_{u_m}$, $l_{p_n}$ and $l_r$ are the numbers of words in $\textbf{U}_m$, $\textbf{P}_n$ and $\textbf{R}$ respectively. Each $\textbf{u}_{m,i}$, $\textbf{p}_{n,j}$ or $ \textbf{r}_k$ is an embedding vector of \emph{d}-dimensions.

  \subsection{Sentence Encoding Layer}
   The context utterances, profile sentences and response candidate are encoded by bidirectional long short-term memories (BiLSTMs) \cite{DBLP:journals/neco/HochreiterS97}. We denote the calculations as follows,
    \begin{align}
    \bar{\textbf{u}}_{m,i} &= \textbf{BiLSTM}(\textbf{U}_m, i), i \in \{1, ..., l_{u_m}\},\\
    \bar{\textbf{p}}_{n,j} &= \textbf{BiLSTM}(\textbf{P}_n, j), j \in \{1, ..., l_{p_n}\},\\
    \bar{\textbf{r}}_{k}   &= \textbf{BiLSTM}(\textbf{R}, k),   k \in \{1, ..., l_{r}\},
    \end{align}
   where $\bar{\textbf{U}}_m = \{\bar{\textbf{u}}_{m,i}\}_{i=1}^{l_{u_m}}$, $\bar{\textbf{P}}_n = \{\bar{\textbf{p}}_{n,j}\}_{j=1}^{l_{p_n}}$ and $\bar{\textbf{R}} = \{\bar{\textbf{r}}_k\}_{j=1}^{l_r}$. The parameters of these three BiLSTMs are shared in our implementation.

  \subsection{Matching Layer}

   The interactions between the context and the response and those between the persona and the response can provide useful matching information for deciding the matching degree between them. Here, the DIM model adopts the same strategy as in the IMN model \cite{DBLP:conf/cikm/GuLL19} which considers the global and bidirectional interactions between two sequences.

   Take the context-response matching as an example. First, the context representation $\bar{\textbf{C}} = \{\bar{\textbf{c}_i}\}_{i=1}^{l_c} $ with $l_c = \sum_{m=1}^{n_c} l_{u_m}$ is formed by concatenating the set of utterance representations $\{\bar{\textbf{U}}_{m}\}_{m=1}^{n_c}$. Then, a soft alignment is performed by computing the attention weight between each tuple \{$\bar{\textbf{c}}_i, \bar{\textbf{r}}_k$\} as
    \begin{align}
    e_{ik} = (\bar{\textbf{c}}_i)^\top \cdot \bar{\textbf{r}}_k.
    \end{align}

   After that, local inference is determined by the attention weights computed above to obtain the local relevance between a context and a response bidirectionally. For a word in the context, its relevant representation carried by the response is identified and composed using $e_{ik}$ as
    \begin{align}
    \tilde{\textbf{c}}_i = \sum_{k=1}^{l_r} \frac{exp(e_{ik})} {\sum_{l=1}^{l_r} exp(e_{il})} \bar{\textbf{r}}_k, i \in \{1, ..., l_c\},
    \end{align}
   where the contents in $\{\bar{\textbf{r}}_k\}_{k=1}^{l_r}$ that are relevant to $\bar{\textbf{c}}_i$ are selected to form $\tilde{\textbf{c}}_i$.
    Then, we define $\tilde{\textbf{C}} = [\tilde{\textbf{c}}_1,...,\tilde{\textbf{c}}_{l_c}]$.
    The same calculation is performed for each word in the response to form its relevant representation carried by the context as
    \begin{align}
    \tilde{\textbf{r}}_k = \sum_{i=1}^{l_c} \frac{exp(e_{ik})} {\sum_{l=1}^{l_c} exp(e_{lk})} \bar{\textbf{c}}_i, k \in \{1, ..., l_r\},
    \label{equ1}
    \end{align}
    and we define $\tilde{\textbf{R}} = [\tilde{\textbf{r}}_1,...,\tilde{\textbf{r}}_{l_r}]$.
    To further enhance the collected information, the differences and element-wise products between \{$\bar{\textbf{C}}, \tilde{\textbf{C}}$\} and between \{$\bar{\textbf{R}}, \tilde{\textbf{R}}$\} are computed, and are then concatenated with the original vectors to obtain the enhanced representations as follows,
    \begin{align}
    \widehat{\textbf{C}} = [\bar{\textbf{C}}; \tilde{\textbf{C}}; \bar{\textbf{C}} - \tilde{\textbf{C}} ;\bar{\textbf{C}} \odot \tilde{\textbf{C}}],\\
    \widehat{\textbf{R}} = [\bar{\textbf{R}}; \tilde{\textbf{R}}; \bar{\textbf{R}} - \tilde{\textbf{R}} ;\bar{\textbf{R}} \odot \tilde{\textbf{R}}].
    \end{align}

    So far we have collected the relevant information between the context and response. The enhanced context representation $\widehat{\textbf{C}}$ is further converted back to matching matrices of separated utterances as $\{ \widehat{\textbf{U}}_m \}_{m=1}^{n_c}$.

    The persona-response matching  is conducted identically to the context-response matching introduced above,
    where the representations of profile sentences $\{\bar{\textbf{P}}_{n}\}_{n=1}^{n_p}$ are used, instead of the representations of context utterances $\{\bar{\textbf{U}}_{m}\}_{m=1}^{n_c}$.
    The results of persona-response matching are denoted as $\{\widehat{\textbf{P}}_n\}_{n=1}^{n_p}$ and $\widehat{\textbf{R}}^*$.

  \subsection{Aggregation Layer}
   The aggregation layer converts the matching matrices of context utterances, profile sentences and response into a final matching feature vector.

   First, each matching matrix $\widehat{\textbf{U}}_m, \widehat{\textbf{R}}, \widehat{\textbf{P}}_n$ and $\widehat{\textbf{R}}^*$ are processed by BiLSTMs as
    \begin{align}
    \hat{\textbf{u}}_{m,i}^{utr} &= \textbf{BiLSTM}(\widehat{\textbf{U}}_m, i), i \in \{1, ..., l_{u_m}\},\\
    \hat{\textbf{r}}_k^{utr}     &= \textbf{BiLSTM}(\widehat{\textbf{R}}, k),   k \in \{1, ..., l_r\},\\
    \hat{\textbf{p}}_{n,j}^{utr} &= \textbf{BiLSTM}(\widehat{\textbf{P}}_n, j), j \in \{1, ..., l_{p_n}\},\\
    \hat{\textbf{r}}_k^{utr*}    &= \textbf{BiLSTM}(\widehat{\textbf{R}}^*, k), k \in \{1, ..., l_r\},
    \end{align}
    where the four BiLSTMs share the same parameters in our implementation.
    Then, the aggregated embeddings are calculated by max pooling and last-hidden-state pooling operations as
    \begin{align}
    \hat{\textbf{u}}_m^{agr} = [\hat{\textbf{u}}_{m,max}^{utr};\hat{\textbf{u}}_{m,l_{u_m}}^{utr}]&, m \in \{1, ..., n_c\},\\
    \hat{\textbf{r}}^{agr}   = [\hat{\textbf{r}}_{max}^{utr};  \hat{\textbf{r}}_{l_r}^{utr}]&,\\
    \hat{\textbf{p}}_n^{agr} = [\hat{\textbf{p}}_{n,max}^{utr};\hat{\textbf{p}}_{n,l_{p_n}}^{utr}]&, n \in \{1, ..., n_p\},\\
    \hat{\textbf{r}}^{agr*} = [\hat{\textbf{r}}_{max}^{utr*};\hat{\textbf{r}}_{l_r}^{utr*}]&.
    \end{align}
    Next, the sequences of $\hat{\textbf{u}}_m^{agr}$ and $\hat{\textbf{p}}_n^{agr}$ are further aggregated to get the embedding vectors for the context and the persona respectively.

    \paragraph{Context aggregation} As the utterances in a context are chronologically ordered, the utterance embeddings $\widehat{\textbf{U}}^{agr}=\{\hat{\textbf{u}}_{m}^{agr}\}_{m=1}^{n_c}$ are sent into another BiLSTM following the chronological order of utterances in the context.
    Combined max pooling and last-hidden-state pooling operations are then performed to obtain the context embeddings as
    \begin{align}
    \hat{\textbf{u}}_m^{ctx} = \textbf{BiLSTM}(\widehat{\textbf{U}}^{agr}, m)&, m \in \{1, ..., n_c\},\\
    \hat{\textbf{c}}^{agr}   = [\hat{\textbf{u}}_{max}^{ctx};\hat{\textbf{u}}_{n_c}^{ctx}]&.
    \end{align}

    \paragraph{Persona aggregation} As the profile sentences in a persona are independent to each other, an attention-based aggregation is designed to derive the persona embeddings as follows,
    \begin{align}
    \alpha_n         &= \textbf{ReLU}(\textbf{w}^\top \cdot \hat{\textbf{p}}_n^{agr} + b),\\
    \hat{\textbf{p}}^{agr} &= \sum_{n=1}^{n_p} \frac{e^{\alpha_n}} {\sum_{k=1}^{n_p} e^{\alpha_k}} \hat{\textbf{p}}_n^{agr},
    \end{align}
    where $\textbf{w}$ and $b$ are parameters need to be estimated during training.

    Last, the final matching feature vector is the concatenation of context, persona and response embeddings as
    \begin{align}
    \textbf{m} = [\hat{\textbf{c}}^{agr};\hat{\textbf{r}}^{agr};\hat{\textbf{p}}^{agr};\hat{\textbf{r}}^{agr*}],
    \end{align}
    where the first two features describe the context-response matching, and the last two describe the persona-response matching.

  \subsection{Prediction Layer}
   The final matching feature vector is then sent into a multi-layer perceptron (MLP) classifier with softmax output. Here, the MLP is designed to predict whether a $(c,p,r)$ triple match appropriately based on the derived matching feature vector. Finally, the MLP returns a probability to denote the matching degree.

\section{Experiments}

  \subsection{Dataset}
   We tested our proposed methods on the PERSONA-CHAT dataset \cite{DBLP:conf/acl/KielaWZDUS18} which contains multi-turn dialogues conditioned on personas. The dataset consists of 8939 complete dialogues for training, 1000 for validation, and 968 for testing. Response selection is performed at every turn of a complete dialogue, which results in 65719 dialogues for training, 7801 for validation, and 7512 for testing in total. Positive responses are true responses from humans and negative ones are randomly sampled. The ratio between positive and negative responses is 1:19 in the training, validation, and testing sets. There are 955 possible personas for training, 100 for validation, and 100 for testing, each consisting of 3 to 5 profile sentences. To make this task more challenging, a version of revised persona descriptions are also provided by rephrasing, generalizing, or specializing the original ones. Since the personas of both speakers in a dialogue are available, the response selection task can be conditioned on the speaker's persona (``self persona") or the dialogue partner's persona (``their persona") respectively.

  \subsection{Evaluation Metrics}
   We used the same evaluation metrics as in the previous work \cite{DBLP:conf/acl/KielaWZDUS18}. Each model aimed to select the best-matched response from available candidates for the given context $c$ and persona $p$. We calculated the recall of the true positive replies, denoted as $\textbf{hits}@1$. In addition, the mean reciprocal rank (\textbf{MRR}) \cite{DBLP:conf/trec/Voorhees99} metric was also adopted to take the rank of the correct response over all candidates into consideration.

  \subsection{Training Details}
   For building the IMN, IMN$_{ctx}$, IMN$_{utr}$ and DIM models, the Adam method \cite{DBLP:journals/corr/KingmaB14} was employed for optimization with a batch size of 16.
   The initial learning rate was 0.001 and was exponentially decayed by 0.96 every 5000 steps.
   Dropout \cite{DBLP:journals/jmlr/SrivastavaHKSS14} with a rate of 0.2 was applied to the word embeddings and all hidden layers.
   A word representation is a concatenation of a 300-dimensional GloVe embedding \cite{DBLP:conf/emnlp/PenningtonSM14}, a 100-dimensional embedding estimated on the training set using the Word2Vec algorithm \cite{DBLP:conf/nips/MikolovSCCD13}, and 150-dimensional character-level embeddings with window sizes \{3, 4, 5\}, each consisting of 50 filters.
   The word embeddings were not updated during training.
   All hidden states of the LSTM have 200 dimensions.
   The MLP at the prediction layer have 256 hidden units with ReLU \cite{DBLP:conf/icml/NairH10} activation.
   The maximum number of characters in a word, that of words in a context utterance, of utterances in a context, and of words in a response
   were set to be 18, 20, 15, and 20, respectively.
   We padded with zeros if the number of utterances in a context was less than 15; otherwise, we kept the last 15 utterances.
   For the IMN$_{ctx}$, IMN$_{utr}$ and the DIM models, the maximum number of words in a profile sentence and that of profile sentences in a persona were set to be 15 and 5, respectively.
   Similarly, we padded with zeros if the number of profile sentences in a persona was less than 5.
   The development set was used to select the best model for testing.

   All code was implemented in the TensorFlow framework \cite{DBLP:conf/osdi/AbadiBCCDDDGIIK16} and is published to help replicate our results~\footnote{https://github.com/JasonForJoy/DIM}.

  \subsection{Experimental Results}

   \begin{table}[t]
     \centering
     \begin{tabular}{l|c|c}
      \toprule
                        & $\textbf{hits}@1$ & \textbf{MRR} \\
      \hline
       IR baseline      & 21.4 & -      \\
       Starspace        & 31.8 & -      \\
       Profile          & 31.8 & -      \\
       KV Profile       & 34.9 & -      \\
      \hline
       IMN              & 63.8 & 75.8   \\
      \bottomrule
      \end{tabular}
      \caption{Evaluation results of the IMN model and previous methods on PERSONA-CHAT dataset without using personas. All the results except ours are copied from \citet{DBLP:conf/acl/KielaWZDUS18}.}
      \label{tab2}
   \end{table}

   \begin{table*}[!hbt]
     \small
     \centering
     \setlength{\tabcolsep}{1.3mm}{
     \begin{tabular}{l|c|c|c|c|c|c|c|c}
      \toprule
                        & \multicolumn{4}{c|}{Self Persona} & \multicolumn{4}{c}{Their Persona} \\
      \hline
                        & \multicolumn{2}{c|}{Original} & \multicolumn{2}{c|}{Revised} & \multicolumn{2}{c|}{Original} & \multicolumn{2}{c}{Revised} \\
      \hline
                        & $\textbf{hits}@1$ & \textbf{MRR} & $\textbf{hits}@1$ & \textbf{MRR} & $\textbf{hits}@1$ & \textbf{MRR} & $\textbf{hits}@1$ & \textbf{MRR} \\
      \hline
       IR baseline\small   & 41.0 (+19.6) & - & 20.7 (-0.7) & - & 18.1 (-3.3) & - & 18.1 (-3.3) & -  \\
       Starspace\small     & 48.1 (+16.3) & - & 32.2 (+0.4) & - & 24.5 (-7.3) & - & 26.1 (-5.7) & -  \\
       Profile\small       & 47.3 (+15.5) & - & 35.4 (+3.6) & - & 28.3 (-3.5) & - & 29.4 (-2.4) & -  \\
       KV Profile\small    & 51.1 (+16.2) & - & 35.1 (+0.2) & - & 29.1 (-5.8) & - & 28.9 (-6.0) & -  \\
       FT-PC\small         & -            & - & 60.7 (-)    & - & -           & - & -    & -  \\
      \hline
       IMN$_{ctx}$         & 64.3 (+0.5)  & 76.2 (+0.4)  & 63.8 (+0.0) & 75.8 (+0.0) & 63.7 (-0.1) & 75.8 (+0.0) & 63.5 (-0.3) & 75.7 (-0.1) \\
       IMN$_{utr}$         & 66.7 (+2.9)  & 78.1 (+2.3)  & 64.0 (+0.2) & 76.0 (+0.2) & 63.9 (+0.1) & 75.9 (+0.1) & 63.7 (-0.1) & 75.7 (-0.1)\\
       DIM                 & \textbf{78.8 (+15.0)} & \textbf{86.7 (+10.9)} & \textbf{70.7 (+6.9)} & \textbf{81.2 (+5.4)} & \textbf{64.0 (+0.2)} & \textbf{76.1 (+0.3)} & \textbf{63.9 (+0.1)} & \textbf{76.0 (+0.2)}  \\
      \bottomrule
      \end{tabular}}
      \caption{Performance of the proposed and previous methods on the PERSONA-CHAT under various persona configurations.
      The meanings of ``Self Persona", ``Their Persona", ``Original", and ``revised" can be found in Section 6.1.
      All results except ours are copied from \citet{DBLP:conf/acl/KielaWZDUS18,DBLP:conf/emnlp/MazareHRB18}. Numbers in parentheses indicate the gains or losses after adding the persona conditions.}
      \label{tab3}
   \end{table*}

   Table~\ref{tab2} presents the evaluation results of our reproduced IMN model \cite{DBLP:conf/cikm/GuLL19} and previous methods on PERSONA-CHAT dataset without using personas. It can be seen that the IMN model outperformed other models on this dataset by a margin larger than 28.9\% in terms of $\textbf{hits}@1$. As introduced above, our proposed models for personalized response selection were all built on IMN.

   Table~\ref{tab3} presents the evaluation results of our proposed and previous methods on PERSONA-CHAT under various persona configurations. The t-test shows that the differences between our proposed models, i.e., IMN$_{utt}$ and DIM,  and the baseline model, i.e. IMN$_{ctx}$, were both statistically significant with \emph{p}-value $<$ 0.01. We can see that the fine-grained persona fusion at the utterance level rendered a $\textbf{hits}@1$ improvement of 2.4\%  and an \textbf{MRR} improvement of 1.9\% by comparing IMN$_{ctx}$ and IMN$_{utr}$ conditioned on original self personas. The DIM model outperformed its baseline IMN$_{ctx}$ by a margin of 14.5\% in terms of $\textbf{hits}@1$ and 10.5\% in terms of \textbf{MRR}. Compared with the FT-PC model \cite{DBLP:conf/emnlp/MazareHRB18} which was first pretrained using a large-scale corpus and then fine-tuned on the PERSONA-CHAT dataset, the DIM model outperformed it by a margin of 10.0\% in terms of $\textbf{hits}@1$ conditioned on revised self personas. Another advantage of DIM is that it was trained in an end-to-end mode without pretraining and using any external knowledge. Lastly, the DIM model outperforms previous models by margins larger than 27.7\% in terms of $\textbf{hits}@1$
   conditioned on original self personas.

   \paragraph{Improvement of Using Personas}
   Examining the numbers which indicate the gains or losses after adding persona conditions in Table \ref{tab3}, we can see that the context-level persona fusion improves the performance of previous models significantly when original self personas are used. However, the gain achieved by the IMN$_{ctx}$ model is limited. One possible reason is that the IMN model performs attention-based interactions between the context and the response in order to get their matching feature for response selection.
   Thus, the context embeddings shown in Fig. \ref{fig3}(a) contained the information from both the context and the response, which may be inappropriate for the following context-level persona fusion shown in Eq.~(\ref{equ2}).
   The improvement achieved by the DIM model is much higher because it adopts a dual matching framework to address this issue.

   \paragraph{Original vs. Revised}
   Comparing with using original personas, it is more difficult for the models conditioned on the revised personas to extract useful persona information, as shown by the limited improvement achieved by the previous models shown in Table \ref{tab3}. One possible reason is that there are fewer shared words between the response and the persona revised by rephrasing, generalizing, or specializing, which increases the difficulty of understanding the persona and its relationships with the response. For example, it is easier for models to judge the matching degree between the original profile ``\emph{Autumn is my favorite \underline{season}.}" and the response ``\emph{This is my favorite time of the year \underline{season} wise.}" than between the revised profile ``\emph{I love watching the leaves change colors.}" and the response. On the contrary, our proposed DIM model still obtains a  $\textbf{hits}@1$ improvement of  6.9\%  and an \textbf{MRR} improvement of 5.4\% when conditioned on the revised self personas, which can be attributed to the direct and interactive persona-response matching used in this model.

   \paragraph{Self vs. Their} As shown in Table \ref{tab3}, no significant gains can be obtained when the models are conditioned on the personas of dialogue partners. Note that there are no significant performance losses with our proposed methods, while the losses of previous models are 2.4\% to 7.3\% in terms of $\textbf{hits}@1$.

\section{Analysis}

  \subsection{Ablations}

  \begin{table}[!hbt]
     \centering
     \begin{tabular}{l|c|c}
      \toprule
                        & $\textbf{hits}@1$ & \textbf{MRR} \\
      \hline
       DIM              & 78.8 & 86.7  \\
       \ \ - persona    & 63.8 & 75.8  \\
       \ \ - context    & 48.8 & 60.9  \\

      \bottomrule
      \end{tabular}
      \caption{Ablation tests of removing either persona-response matching or context-response matching in the DIM model conditioned on original self personas.}
      \label{tab4}
   \end{table}

   To demonstrate the importance of the dual matching framework followed by our proposed DIM model, ablation tests were performed using the original self personas and the results are shown in Table~\ref{tab4}. We can see that both the persona-response matching and the context-response matching contribute to the performance of the DIM model. It is reasonable that the context-persona matching is more important because contexts provide the fundamental semantic descriptions for response selection. On the other hand, the single persona-response matching can also achieve a $\textbf{hits}@1$ of 48.8\% and an $\textbf{MRR}$ of 60.9\%, which shows the usefulness of utilizing persona information to select the best-matched response.

  \subsection{Interactive Matching in DIM}

\begin{figure}
    \centering
    \subfigure[Response-to-context attention weights]{
    \includegraphics[width=7.5cm]{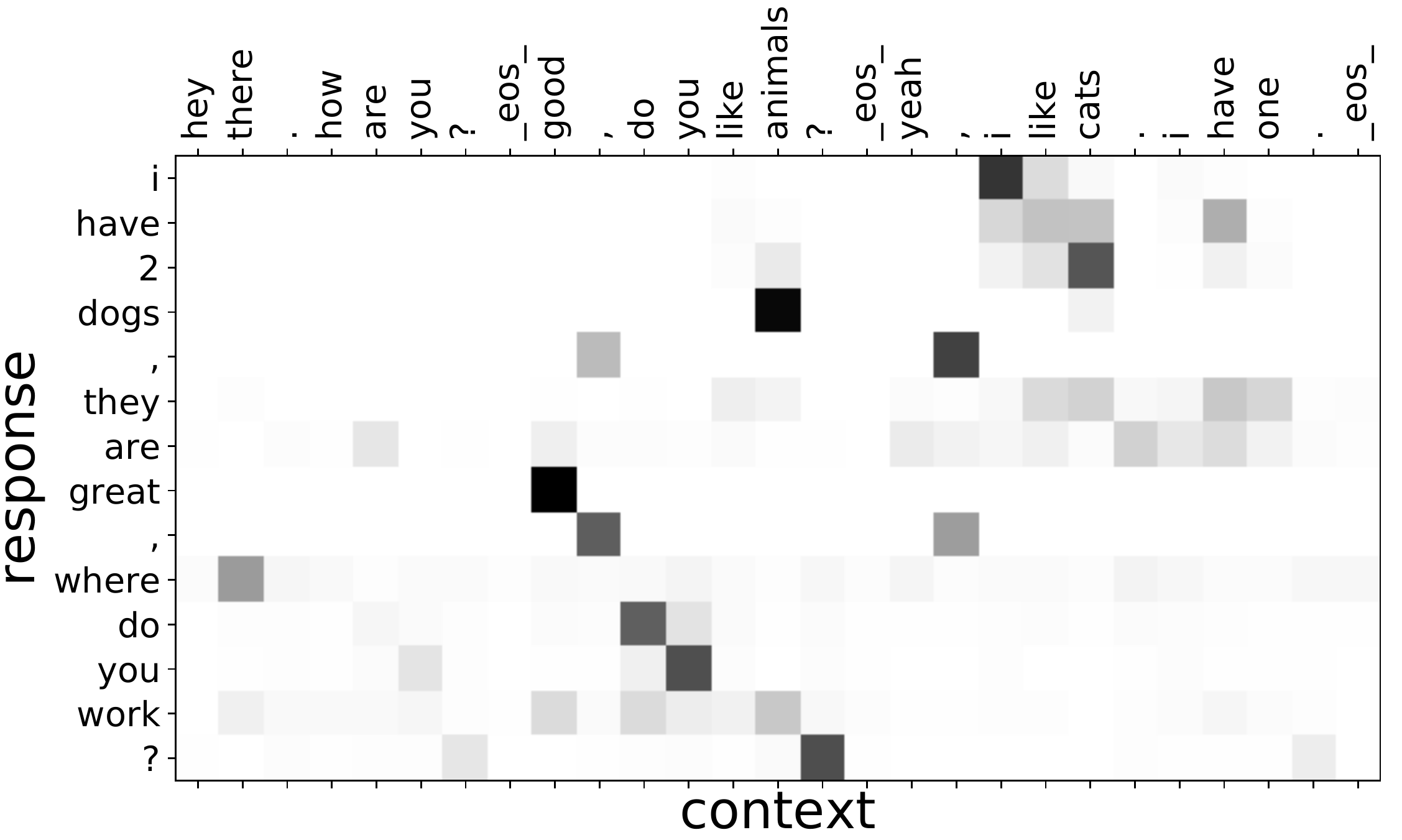}}
    \subfigure[Response-to-persona attention weights]{
    \includegraphics[width=7.5cm]{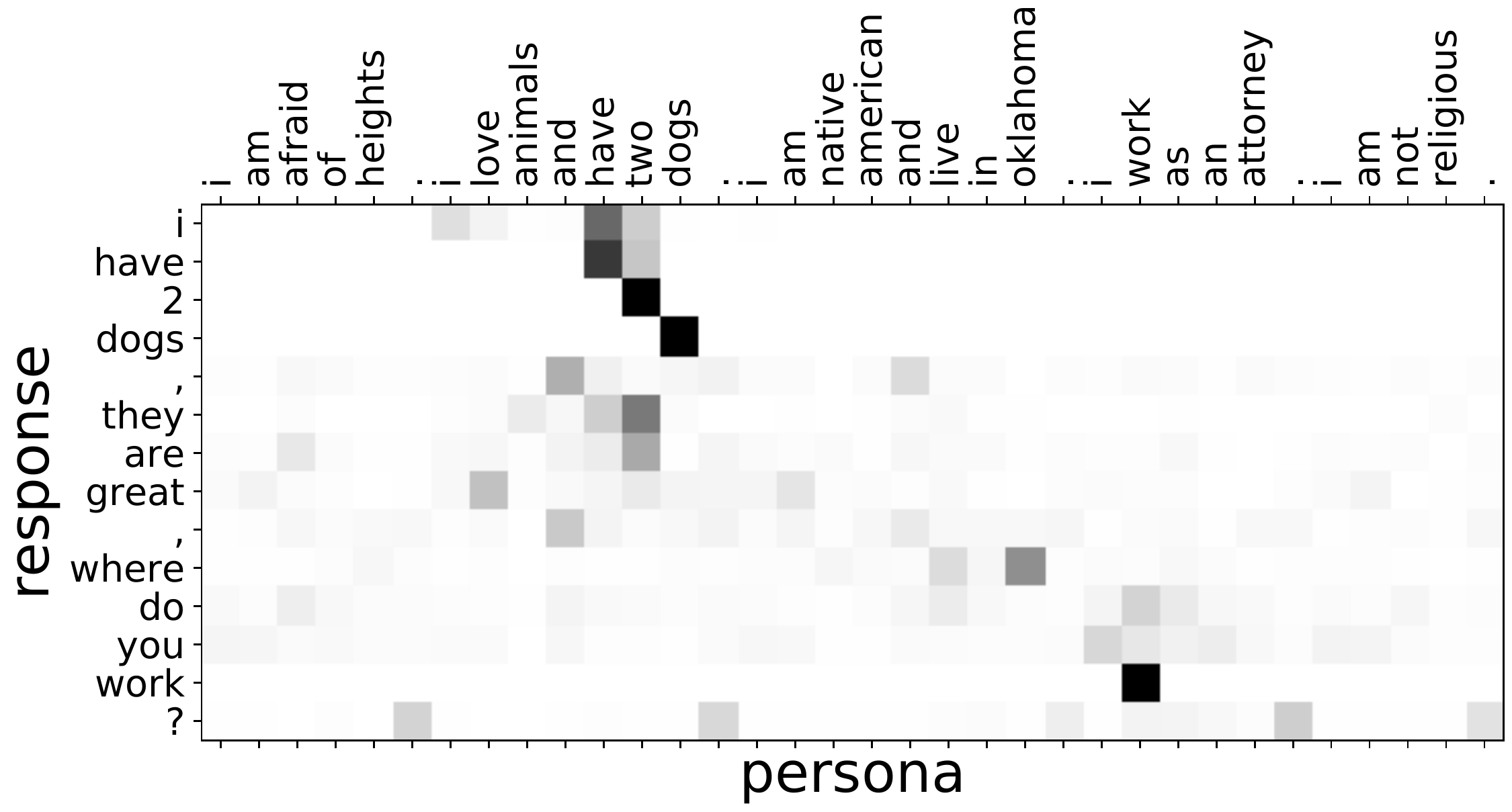}}
    \caption{Visualizations of (a) response-to-context or (b) response-to-persona attention weights at the matching layer for a test sample. The darker units correspond to larger values.}
    \label{fig2}
   \end{figure}

    In order to investigate the effectiveness of the interaction matching between the context and the response and that between the persona and the response in the DIM model, a case study was conducted by visualizing the response-to-context and response-to-persona attention weights used in Eq.~(\ref{equ1}). The results are shown in Fig.~\ref{fig2}. We can see that some important words such as ``\emph{dogs}" in the response selected its relevant words such as ``\emph{animals}" in the context to derive the context-response matching features. Some important profile texts such as ``\emph{I love animals and have two dogs.}" also obtained large attention weights for getting the persona-response matching features. This experimental result confirms our formulation of the task of personalized response selection as a dual matching problem.

  \subsection{Transfer Test}

   \begin{table}[!hbt]
     \centering
     \begin{tabular}{l|c|c}
      \toprule
      \diagbox{Train}{Test}  & Original & Revised \\
      \hline
       Original              & 78.8     & 66.3  \\
       Revised               & 77.6     & 70.7  \\
      \bottomrule
      \end{tabular}
      \caption{$\textbf{hits}@1$ results of transfer tests on the DIM model.}
      \label{tab5}
   \end{table}

   Transfer tests were conducted by training and evaluating the DIM model using mismatched types of personas. The results are reported in Table~\ref{tab5}. It shows that the DIM model achieved a better performance when testing on the same type of personas as training. Meanwhile, the model trained on the revised personas and tested on the original personas achieved less loss than the ones trained on the original personas and tested on the revised personas, which shows that the revised personas can provide a better generalization ability to the DIM model than the original ones.

\section{Conclusions}
  In this paper, we formulate the task of personalized response selection as a dual matching problem to search for a response that can properly match the given context and persona simultaneously. A new model named Dually Interactive Matching Network (DIM) is proposed, which performs the interaction matching between the context and response as well as between the persona and response in parallel, in order to derive the final matching features for personalized response selection. Experimental results show that DIM improves over the IMN models with context-level or utterance-level persona fusion, outperforming previous methods and achieving a new state-of-the-art performance on the PERSONA-CHAT dataset. In the future, we will explore models to make better use of dialogue partners' persona for response selection.

\section*{Acknowledgments}
  We thank the anonymous reviewers for their valuable comments.

\bibliography{emnlp-ijcnlp-2019}
\bibliographystyle{acl_natbib}

\end{document}